\documentclass{article}

\usepackage{microtype}
\usepackage{graphicx}
\usepackage{booktabs} 
\usepackage{multicol}
\usepackage{multirow}
\usepackage{caption}
\usepackage{subcaption}
\usepackage[final]{changes}

\usepackage{hyperref}

\usepackage[accepted]{icml2024}

\usepackage{amsmath}
\usepackage{amssymb}
\usepackage{mathtools}
\usepackage{amsthm}
\usepackage{algorithmic}

\usepackage[capitalize,noabbrev]{cleveref}

\theoremstyle{plain}
\newtheorem{theorem}{Theorem}[section]
\newtheorem{proposition}[theorem]{Proposition}

\theoremstyle{definition}

\theoremstyle{remark}

\DeclareMathOperator*{\argmin}{arg\,min}

\newcommand{\eg}{\emph{e.g.}}
\newcommand{\ie}{\emph{i.e.}}
\newcommand{\etal}{\emph{et al.}}

\makeatletter
\newcommand*\bigcdot{\mathpalette\bigcdot@{.5}}
\newcommand*\bigcdot@[2]{\mathbin{\vcenter{\hbox{\scalebox{#2}{$\m@th#1\bullet$}}}}}
\makeatother

\icmltitlerunning{Low-Rank Similarity Mining}

\begin{document}

\twocolumn[
\icmltitle{Low-Rank Similarity Mining for Multimodal Dataset Distillation}


\icmlsetsymbol{equal}{*}

\begin{icmlauthorlist}
\icmlauthor{Yue Xu}{sjtu}
\icmlauthor{Zhilin Lin}{sjtu}
\icmlauthor{Yusong Qiu}{sjtu}
\icmlauthor{Cewu Lu}{sjtu}
\icmlauthor{Yong-Lu Li}{sjtu}
\end{icmlauthorlist}

\icmlaffiliation{sjtu}{Shanghai Jiao Tong University}
\icmlcorrespondingauthor{Yong-Lu Li}{yonglu\_li@sjtu.edu.cn}

\icmlkeywords{Dataset Distillation, Visual-Language Learning, Low Rank}

\vskip 0.3in
]



\printAffiliationsAndNotice{}  

\begin{abstract}
Though dataset distillation has witnessed rapid development in recent years, the distillation of multimodal data, \eg, image-text pairs, poses unique and under-explored challenges. 
Unlike unimodal data, image-text contrastive learning (ITC) data lack inherent categorization and should instead place greater emphasis on modality correspondence. 
In this work, we propose \textbf{Lo}w-\textbf{R}ank \textbf{S}imilarity Mining (\textbf{LoRS}) for multimodal dataset distillation, that concurrently distills a ground truth similarity matrix with image-text pairs, and leverages low-rank factorization for efficiency and scalability. 
The proposed approach brings significant improvement to the existing algorithms, marking a significant contribution to the field of visual-language dataset distillation.
We advocate adopting LoRS as a foundational synthetic data setup for image-text dataset distillation. 
\textbf{Our code is available at \url{https://github.com/silicx/LoRS_Distill}}.
\end{abstract}

\section{Introduction}
\label{sec:intro}

Dataset distillation synthesizes a smaller and more compact dataset while retaining its essential information and model training performance.
It becomes noteworthy in machine learning due to its high compression ratio, especially in the context of large-scale models and data.
However, current algorithms are limited in the image domain and very few works involve other uni-modality such as text~\cite{li2021data-TEXT}, video~\cite{wang2023dancing-VIDEO} or graph~\cite{xu2023kernel-GRAPH} data. Since vision-language pre-training models (VLP)~\cite{clip, blip} and multimodal large language models (MLLM)~\cite{blip2, llava} become dominant, we direct our attention towards the paired image-text data. As the foundation of VLP, we focus on the image-text contrastive learning (ITC) data and aim for an effective image-text dataset distillation which could potentially enhance the efficiency and boost the study of multimodal models.

However, the distillation of image-text pairs is much more challenging than unimodal data:
(1) The algorithm should not only compress each modality individually but should also correctly learn the \textit{correspondence} between the modalities;
(2) The unimodal data have categories and are distributed in clusters; but the image-text pair data is not categorized and distributed sparsely, which could lead to high sample variance for dataset distillation. As indicated by previous work~\cite{KFS}, this fails existing algorithms like gradient matching~\cite{DC} and distribution matching~\cite{DM}. 
\added{
While the first work on image-text dataset distillation~\citep{vl-distill} achieves non-trivial performance with vanilla MTT~\cite{MTT}, it lacks specific adaption and exploitation of the image-text data.
}
Therefore, we propose to emphasize learning the modality correspondence rather than summarizing the data patterns for each category.

\begin{figure}[t]
\begin{center}
\centerline{\includegraphics[width=0.97\columnwidth]{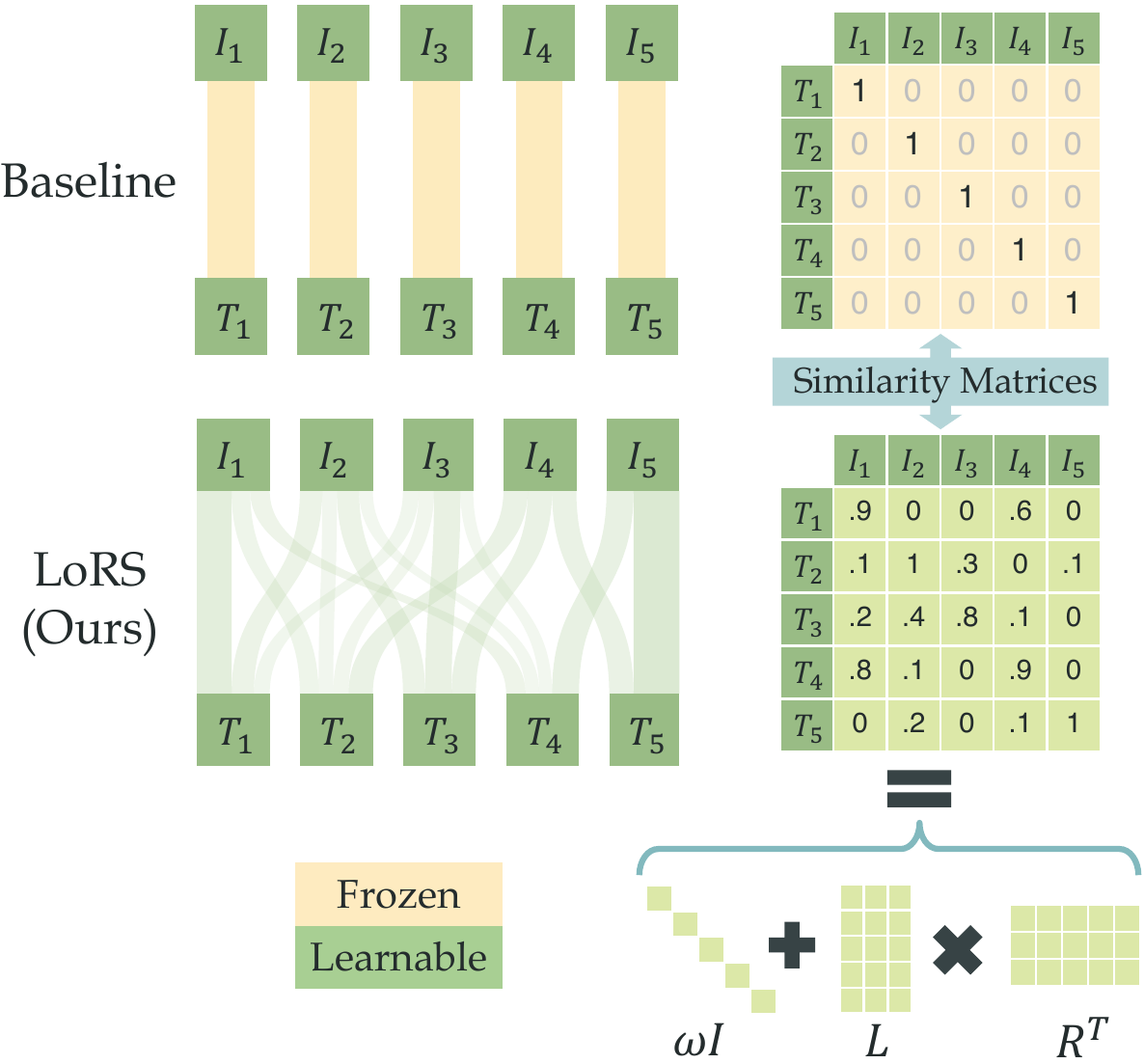}}
\caption{Vanilla dataset distillation could be adapted to image-text data but is limited by the fixed data pairing (``Baseline''). We propose similarity mining which simultaneously distills the ground truth similarity matrix, together with low-rank optimization for a fair data parameter size (\textbf{LoRS}). ($I_i$ = $i^\text{th}$ image, $T_i$ = $i^\text{th}$ text).}
\vspace{-5px}
\label{fig:teaser}
\vspace{-15px}
\end{center}
\end{figure}

As shown in Fig.~\ref{fig:teaser}, vanilla dataset distillation algorithms exploit \textit{fixed} image-text correspondence. To enhance the information density and bring more flexibility, we propose simultaneously learning the image-text similarity during dataset distillation as auxiliary information, \ie, \textbf{Similarity Mining}. 
The distilled similarity matrix could be exploited in \textit{any} image-text contrastive learning models, with only slight modifications to the contrastive loss functions.
This method extends the vanilla distillation method to learn the full correspondence between synthetic image and text, which could be roughly considered as extending the $N$ image-text pairs to $N^2$ paired data. 
Thus, we enrich the information of the synthetic data with little model overhead. 
We advocate adopting similarity mining as a fundamental algorithm setting in image-text dataset distillation.

To support the rationality and feasibility, we justify the similarity mining from the model learning perspectives:
(1) \textbf{False negative mining}: the vanilla ITC models (\eg, CLIP~\cite{clip}) assume the samples in each batch are distinct so it uses the identity matrix as Ground Truth (GT) similarity matrix (only the sample itself is positive and other samples are all negative), but sometimes potential similarity between batch samples exists~\cite{CWCL}, and similarity mining could find these samples and automatically fix the false loss penalty.
(2) \textbf{Flexible contrastive learning anchors}: ITC can be regarded as the attraction and repulsion between the feature embedding and the anchor points. Similarity mining gives the flexibility to distinctly weigh the anchors so that some anchors can be equivalently merged without changing the learning dynamics, which will largely enhance the compression rate for dataset distillation.
These will be detailed in Sec.~\ref{sec:justify}.

The similarity mining works well when the data size $\tilde{M}$ is small.
However, when the synthetic data size scales up, the parameter size of the similarity matrix could explode quadratically.
\eg, when $\tilde{M}=100$, the similarity matrix is smaller than the one single image; but when $\tilde{M}$ grows to 10,000, the matrix is larger than the total image and text storage. 
Thus, we find that the similarity matrix is \textit{sparse} in the non-diagonal area. For memory efficiency and to facilitate the learning of the similarity matrix, we exploit low-rank factorization for the similarity matrix.
The similarity matrix is decomposed: $S=\omega I+\frac{\alpha}{r}L R^\top$, where the low-rank components $L, R$ are $\tilde{M}\times r$ sized and reduce the space complexity to $O(\tilde{M})$.
Overall, we propose \textbf{Lo}w \textbf{R}ank \textbf{S}imilarity Mining (\textbf{LoRS}) for image-text dataset distillation. Experiments show that our methods significantly enhance the distilled data performance and compression ratio.

The contribution of this work involves:
\textbf{(1)} For image-text dataset distillation, we propose a new paradigm to learn the similarity matrix as a part of the synthetic data, which is well justified from the ITC training perspective.
\textbf{(2)} We propose a novel and feasible implementation of similarity mining incorporating low-rank factorization.
\textbf{(3)} Our method significantly outperforms the baseline methods with the same or smaller storage burden.

\section{Related Work}

\subsection{Dataset Distillation}

Dataset distillation (DD) aims to synthesize a small-scale dataset from a large-scale dataset, which can replace the original dataset in training while maintaining performance. Existing algorithms can be classified as:
(1)	\textbf{Meta-Model Matching}. Optimizing the empirical loss on the full dataset and keeping the transferability of distilled data. Following the first work of DD~\cite{DD}, many approaches have been proposed. KIP~\cite{KIP} integrates ridge regression to reduce computational complexity and further extend to an infinite wide network~\cite{KIP2}. RFAD~\cite{RFAD} uses the Neural Network Gaussian Process kernel substitute in KIP. FRePo~\cite{FRePo} divides the network into a feature extractor and a classifier for optimization. RCIG~\cite{RCIG} exploits implicit gradient to compute meta-gradient.
(2) \textbf{Gradient-based}. DC~\cite{DC} aligns the training gradients of real and synthetic data. IDC~\cite{IDC} improves DC by storing synthetic data in lower resolutions. MTT~\cite{MTT} matches the parameters after multi-step training, which can be regarded as long-term gradient matching. TESLA~\cite{tesla} reduces the memory consumption of MTT. Shin~\etal matches the loss sharpness of real and synthetic data, which is similar to the gradient.
(3) \textbf{Feature-based}. DM~\cite{DM} matches the distribution between real data and synthetic data, while CAFE~\cite{CAFE} introduces layer-wise alignment of features. IDM~\cite{IDM} further enhances DM with regularizer and model queue.
(4) \textbf{Factorization Methods}. These methods decompose the data into bases and hallucinators, which can reduce the storage burden and increase the diversity of synthetic data. HaBa~\cite{haba} employs convolutional network hallucinators, while LinBa~\cite{linba} uses linear ones. KFS~\cite{KFS} provides efficient sharing of information between generated examples and a better trade-off between compression ratio and quality. Frequency domain factorization has also been adopted~\cite{frequency}.

Many other methods go beyond these categories and introduce innovations to DD.
To optimize the existing methods, some works focus on data or model augmentation~\cite{DSA, ModelAug} for generalization of DD, while some exploit sample selection for efficient DD~\cite{DREAM,GoldFromOres,yoco} or to extend the application~\cite{liu2023slimmable,yoco}.
Generative models are used as synthetic image generators~\cite{IT-GAN,Generative}.
SRe2L~\cite{sre2l} proposes a 3-stage learning paradigm that is more efficient for large datasets.
Bayesian inference also could be adopted for dataset distillation~\cite{manousakas2020bayesian, kim2022divergence, tiwary2023constructing}.
\added{
Wu~\etal propose the first work on image-text dataset distillation~\citep{vl-distill} and achieve decent performance, by matching the trajectories of both image and text encoders. However, it does not make specific adaptations to the image-text data.
}

\subsection{Image-text Contrastive Learning}

Image-text contrastive learning is a crucial foundation of multimodal learning.
CLIP~\cite{clip} first adopts image-text contrastive learning which aligns the features obtained from encoders of different modalities. The model is trained on large-scale data to achieve \textit{scale effect} and open-vocabulary transferability. ALIGN~\cite{ALIGN} and Flava~\cite{FLAVA} were among the first to present work on comparative learning. CHiLS~\cite{chils} explored richer embedding with label hierarchy. FILIP~\cite{yao2021filip} explored the toke-wise similarity between two modalities. ALBEF~\cite{albef} and CoCa~\cite{yu2022coca} focused on cross-modal attention. BLIP~\cite{blip} and BLIP2~\cite{blip2} made a combination of methods for multimodal learning and performed well.
There is also some recent work focusing on soft labels in CLIP-like models. SoftCLIP~\cite{softclip} achieved soft cross-modal alignment by generating intra-modal similarity. \citet{andonian2022robust} used progressive self-distillation to learn robust models from noisy data.

\section{Methodology}

\subsection{Preliminary}

We first formulate the image-text contrastive learning (ITC).
Based on an image-text pair dataset $\mathcal{D}=(\mathcal{X}, \mathcal{Y})$ with $M$ paired images $\mathcal{X}=\{x_i\}_M$ and texts $\mathcal{Y}=\{y_i\}_M$, an ITC model (\eg, CLIP) consists of an image encoder $u_i=f_V(x_i)$ and a text encoder $v_i=f_T(y_i)$. The model enables cross-modal retrieval with the cosine similarity metric $\hat{s}_{ij}=\cos\langle u_i, v_j\rangle$. 
The model is trained with contrastive learning such as InfoNCE~\cite{InfoNCE} loss:
\begin{equation}
\begin{aligned}
  & \mathcal{L}_\text{NCE}(\mathcal{B}) = - \frac1m\sum_{i=1}^m \log(P^V_{ii})+\log(P^T_{ii}), \\
  & P^V_{ij} = \frac{\exp(\hat{s}_{ij}/\tau)}{\sum_{k}^{m}\exp(\hat{s}_{ik}/\tau)},~P^T_{ij} = \frac{\exp(\hat{s}_{ij}/\tau)}{\sum_{k}^{m}\exp(\hat{s}_{kj}/\tau)},
\end{aligned}
\end{equation}
where $\mathcal{B}\subset \mathcal{D}$ is a data batch with size $m$.
$P^V_{ij}$ and $P^T_{ij}$ are softmax probability of the estimated similarity $\hat{s}_{ij}$ and $\tau$ is a temperature factor.
InfoNCE loss assumes that for each image $x_i$, only the paired text $y_i$ is the positive anchor, and the other texts $y_k, k\neq i$ are negative. So it is aligning the estimated similarity matrix $\hat{S}=\{\hat{s}_{ij}\}$ to an identity GT similarity matrix $I$.

\begin{figure}[t]
\begin{center}
\centerline{\includegraphics[width=\columnwidth]{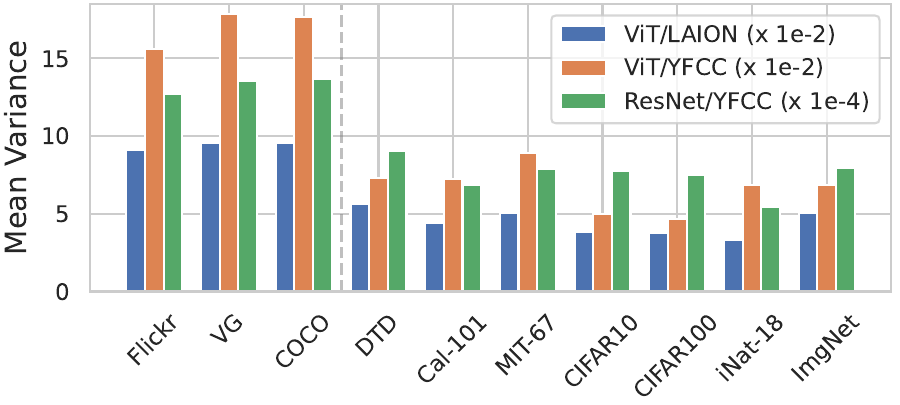}}
\vspace{-3mm}
\caption{The image feature variance on different datasets. \added{We adopt CLIP encoders with ResNet or ViT, pretrained on LAION or YFCC datasets. Three datasets at left: image-text datasets; seven at right: classification datasets.}}
\label{fig:data-variance}
\end{center}
\vspace{-8px}
\end{figure}

The image-text data distillation is to \textbf{learn} a smaller synthetic dataset $\tilde{\mathcal{D}}=( \tilde{\mathcal{X}}, \tilde{\mathcal{Y}} )$, \ie, a data distillation algorithm would regard $\tilde{\mathcal{X}}, \tilde{\mathcal{Y}}$ as learnable parameters and optimize them based on the knowledge from real dataset $\mathcal{D}$.
Most existing dataset distillation algorithms could be seamlessly applied to image-text data. However, Tab.~\ref{tab:dc-dm-result} indicates that the short-term matching algorithms such as gradient matching~\cite{DC} and distribution matching~\cite{DM} fail. This is due to the large data variance of image-text data since no category exists (Fig.~\ref{fig:data-variance}, \added{details please refer to Appendix Sec.~\ref{sec:variance}}). As discussed by ~\cite{KFS}, these methods will suffer from the high batch variance, so they fail on the image-text data.
Therefore, we exploit long-term matching algorithms like MTT~\cite{MTT}.
It matches the model parameters after being trained on the real or synthetic data for multiple training steps so that the batch variance can be alleviated due to multi-step training.

\begin{table}[t]
\caption{Retrieval performance of different distillation algorithms. 200 pairs are synthesized on Flickr30k~\cite{Flickr}. Details of the metrics please refer to Sec~\ref{sec:data-metric}.}
\label{tab:dc-dm-result}
\begin{center}
\begin{small}
\begin{sc}
\resizebox{\linewidth}{!}{
    \begin{tabular}{l|ccc}
        \toprule
        & \shortstack{DC\\\cite{DC}} & \shortstack{DM\\\cite{DM}} & \shortstack{TESLA\\\cite{tesla}} \\
        \midrule
        IR@1  & 0.1$\pm$0.0 & 0.6$\pm$0.1 & \bf  8.6$\pm$0.3 \\
        IR@5  & 0.4$\pm$0.1 & 2.1$\pm$0.2 & \bf 25.3$\pm$0.2 \\
        IR@10 & 1.0$\pm$0.1 & 3.7$\pm$0.4 & \bf 36.6$\pm$0.3 \\
        \midrule
        TR@1  & 0.3$\pm$0.0 & 1.1$\pm$0.3 & \bf 14.5$\pm$0.5 \\
        TR@5  & 0.6$\pm$0.0 & 4.5$\pm$0.1 & \bf 38.7$\pm$0.5 \\
        TR@10 & 1.0$\pm$0.2 & 7.1$\pm$0.1 & \bf 53.4$\pm$0.5 \\
        \bottomrule
    \end{tabular}
}
\vspace{-6px}
\end{sc}
\end{small}
\end{center}
\end{table}

In detail, assume the ITC model $\{f_V,f_T\}$ is parameterized by $\theta=(\theta_V,\theta_T)$. Starting from any initial parameter $\theta_0$, MTT algorithms would train the ITC model on real data for $T$ steps to $\theta_T$, and train on synthetic data $\tilde{\mathcal{X}}, \tilde{\mathcal{Y}}$ for $t$ steps to $\tilde{\theta}_t$.
The synthetic data is optimized by minimizing the long-term parameter matching loss:
\begin{equation}
  \tilde{\mathcal{X}}, \tilde{\mathcal{Y}} = \argmin_{\tilde{\mathcal{X}}, \tilde{\mathcal{Y}}} \frac{
    \|\tilde{\theta}_t - \theta_T\|^2
  }{
    \|\theta_0 - \theta_T\|^2
  }.
\end{equation}
$\tilde{\theta}_t$ is the function of $\tilde{\mathcal{X}}, \tilde{\mathcal{Y}}$ so the synthetic data will receive the gradients for optimization.
In practice, we use TESLA~\cite{tesla} to reduce the memory consumption of the MTT algorithm.

\subsection{Similarity Mining for Image-Text Distillation}

Different from the uni-modal dataset distillation, for the image-text distillation, we should put more emphasis on learning the \textit{correspondence} between the modalities other than the compression of each single modality.
To enhance the information density of image-text data, we propose to simultaneously learn the GT similarity matrix, namely \textbf{similarity mining}. For $M$ image-text pairs, the traditional ITC model assumes identity GT similarity, \ie, only $i^\text{th}$ image and $i^\text{th}$ text are paired and others are all negative pairs. 
However, similarity mining assumes each combination of the data $x_i,y_j$ is a pair but with different GT similarity $s_{ij}$, therefore we have $M\times M$ valid data pairs without increasing the data scale.
The traditional ITC model is a special case when similarity matrix $S=I$.
This intuitively enhances the correspondence information within the same number of image-text data.

Since InfoNCE focuses on the positive pairs, we shall extend the contrastive loss to fit a continuous GT similarity matrix to enable similarity mining.
Given GT similarity $S=\{s_{ij}\}$, we propose the following implementations:

\noindent{\bf Extended InfoNCE (eNCE)}.
We could extend InfoNCE to consider both positives and negatives:
\begin{equation}
\begin{aligned}
    \mathcal{L}_\text{eNCE}(\mathcal{B}, S) =
    & - \frac1m\sum_{i=1}^m\sum_{j=1}^m s_{ij}\left(\log(P_{ij}^V)+\log(P_{ij}^T)\right).
\end{aligned}
\end{equation}
The eNCE resembles CWCL~\cite{CWCL} but without normalization. Note that the loss degrades to InfoNCE when GT $S=I$.

\noindent{\bf Binary Cross Entropy (BCE)}.
The multi-label binary cross-entropy loss is suitable for the continuous similarity:
\begin{equation}
\begin{aligned}
  \mathcal{L}_\text{BCE}(\mathcal{B}, S) &=
    \frac1m\sum_{i=1}^m\sum_{j=1}^m \ell\left(s_{ij}, \sigma(\hat{s}_{ij}/\tau)\right),
\end{aligned}
\end{equation}
where $\sigma$ is the sigmoid function and $\ell(y,p)=-y\log(p)-(1-y)\log(1-p)$.

\noindent{\bf Weighted Binary Cross Entropy (wBCE)}.
Unlike InfoNCE, the BCE loss produces a large penalty for negative pairs. Since there are significantly more negative pairs than positives, we propose a weighted version of BCE by separately averaging the losses for positives and negatives:
\begin{equation}
\begin{aligned}
    \mathcal{L}_\text{wBCE}(\mathcal{B}, S) =
    & \frac1{|\{s_{ij}>\beta\}|}\sum_{i,j:s_{ij}>\beta} \ell\left(s_{ij}, \sigma(\hat{s}_{ij}/\tau)\right) \\
    + & \frac1{|\{s_{ij}\leq\beta\}|}\sum_{i,j:s_{ij}\leq\beta} \ell\left(s_{ij}, \sigma(\hat{s}_{ij}/\tau)\right),
\end{aligned}
\end{equation}
where $\beta$ is the positive/negative threshold and set to 0.5.

More specifically, based on these continuous contrastive losses, the image-text distillation with similarity mining is to learn an augmented synthetic dataset $\tilde{\mathcal{D}}=(\tilde{\mathcal{X}}, \tilde{\mathcal{Y}}, \tilde{S})$, where $\tilde{S}=\{\tilde{s}_{ij}\}$ and $\tilde{s}_{ij}$ is the GT similarity between image $x_i$ and text $y_j$. The synthetic data is learned with MTT loss:
\begin{equation}
  \tilde{\mathcal{X}}, \tilde{\mathcal{Y}}, \tilde{S} = \argmin_{\tilde{\mathcal{X}}, \tilde{\mathcal{Y}}, \tilde{S}} \frac{
    \|\tilde{\theta}_t - \theta_T\|^2
  }{
    \|\theta_0 - \theta_T\|^2
  }.
\end{equation}
To use the distilled data, all the ITC models could be trained with the continuous contrastive losses above on the augmented synthetic dataset (Alg.~\ref{alg:eval}).
Similarity mining could be seen as a data reorganization method like LinBa~\cite{linba}, HaBa~\cite{haba}, so it is \textbf{plug-and-play} for any base distillation algorithm. We will demonstrate its effectiveness both theoretically and empirically, foreseeing much potential for visual language pretraining and multimodal large models. We hope it will become the standard paradigm for image-text dataset distillation.

\subsection{Justification of Similarity Mining}
\label{sec:justify}

The similarity mining technique could be justified from two perspectives:

\begin{figure}[t]
  \begin{center}
  \centerline{\includegraphics[width=\columnwidth]{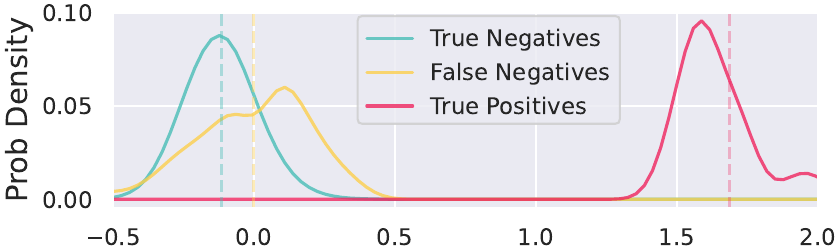}}
  \caption{The histogram of the similarity value learned by similarity mining. False negatives are deliberately constructed and can be found by the algorithm.}
  \vspace{-4px}
  \label{fig:potential-positive}
  \vspace{-8px}
  \end{center}
\end{figure}

\textbf{False negative mining}. Standard ITC models like CLIP assume that the image and text from different samples are negative pairs, which could be violated due to potential same or similar data samples in the noisy internet data.
These potentially associated pairs could be ignorable for large-scale datasets like YFCC100M~\cite{yfcc100m} or LAION~\cite{laion-5b} since there are enough true positives and negatives to draw the representations to the correct position.
However, the small scale of synthetic data leads to low robustness to the false negatives and it requires a more accurate GT similarity.

So the similarity mining paradigm could alleviate the impact of false negatives since it can impose non-zero similarity for \textit{potential} negative pairs.
We conduct a toy experiment on Flickr30k. We initialize 100 synthetic pairs with 50 real data pairs and their replicates so that the $i^\text{th}$ and $(i+1)^\text{th}$ samples will be similar during distillation but regarded as negative pairs. Finally, there are 100 true positive pairs, 100 false negatives, and 9,800 true negatives. We show the normalized histogram of the distilled similarity in Fig.~\ref{fig:potential-positive} and the similarity mining technique does find the false negatives by learning a relatively larger similarity value.

\textbf{Flexible contrastive learning anchors}. We would dig deeper into the image-text contrastive learning by first analyzing the contrastive loss gradients. For conciseness, the following discussion assumes the image and text representations $u_i, v_j$ are normalized, and without loss of generality, we only discuss the gradient on image representation.

\begin{proposition}
\label{prop:NCE}
The gradient of InfoNCE loss wrt. the image representation $u_n$ is:
\begin{equation}
    \frac{\partial \mathcal{L}}{\partial u_n} = \sum_{j=1}^m W_jv_j,~
    W_j=\begin{cases}
    \frac{P^V_{nj} + P^T_{nj} - 2}{m\tau}, & \text{if}~~j=n \\
    \frac{P^V_{nj} + P^T_{nj}}{m\tau},     & \text{if}~~j\neq n \\
    \end{cases}.
\end{equation}
\end{proposition}
The softmax probabilities $P^V_{nj}$ and $P^T_{nj}$ can be interpreted as the relative similarity between $u_n$ and $v_j$.
Given learning rate $\gamma$, the training dynamical system:
\begin{equation}
\label{eq:dynamics}
    \dot{u}_n=-\gamma\frac{\partial \mathcal{L}}{\partial u_n}=\sum_{j=1}^m (-\gamma W_j)v_j.
\end{equation}

\begin{figure}[t]
  \begin{center}
  \centerline{\includegraphics[width=0.85\columnwidth]{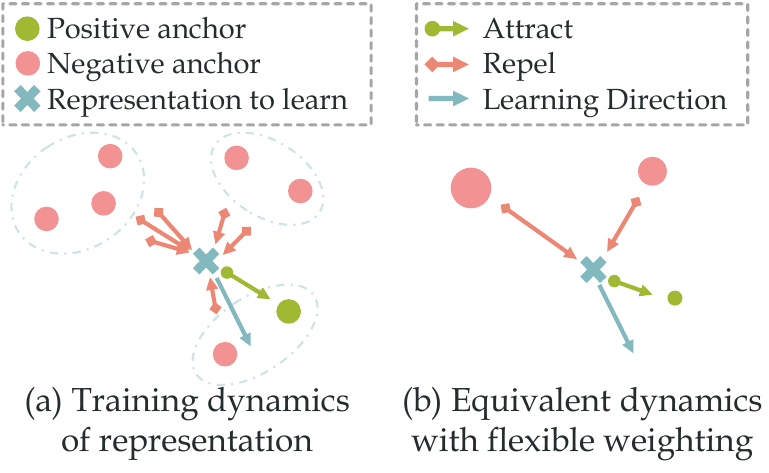}}
  \vspace{-3px}
  \caption{(a) Training dynamical system of a representation: it is attracted or repelled by the anchors. (b) If the anchor is flexibly weighted, dynamics could be equivalent to a system that has fewer components, and similarity mining could offer this flexibility.}
  \label{fig:equivalent-anchor}
  \vspace{-10px}
  \end{center}
\end{figure}

So in a physical analogy, the training dynamics resemble the particle $u_n$ is ``attracted'' or ``repelled'' by other particles $v_j$. The positive or negative \textit{anchor points} $v_j$ span a gradient field which displaces the representation $u_i$ through time.
Eq.~\ref{eq:dynamics} also reveals \textit{approximately} linearity of the dynamics, \ie, the anchors $v_j$ independently affect the representation $u_n$ and the overall dynamics depend on their linear superposition. 
It is not strictly linear since the $P^V$ and $P^T$ are \textit{relative} similarity that involve other $v_k$ components.

Since the system is linear, as shown in Fig.~\ref{fig:equivalent-anchor}, it can be intuitively considered that if the anchor points have distinct weights, then multiple anchor points could be merged as one and the system will be equivalently transformed into a system with fewer components, which reduces the necessary data points and enables a larger compression ratio for dataset distillation.
Here, the similarity mining offers flexibility and induces learnable weights $W_j$:

\begin{proposition}
\label{prop:eNCE}
The gradient of extended InfoNCE loss wrt. the image representation $u_n$ is:
\begin{equation}
    \frac{\partial \mathcal{L}}{\partial u_n} = \sum_{j=1}^m W_jv_j,~W_j=
    \frac{s_{n\bigcdot}P^V_{nj} + s_{\bigcdot j}P^T_{nj} - 2s_{nj}}{m\tau},
\end{equation}
$s_{n\bigcdot}=\sum_{k}^m s_{nk}$ and $s_{\bigcdot j}=\sum_{k}^m s_{kj}$ are marginal similarity.
\end{proposition}

\begin{proposition}
\label{prop:BCE}
The gradient of BCE loss wrt. the image representation $u_n$ is:
\begin{equation}
    \frac{\partial \mathcal{L}}{\partial u_n} = \sum_{j=1}^m W_jv_j,~W_j=
    \frac{ \sigma(\hat{s}_{nj}/\tau) - s_{nj} }{m\tau}.
\end{equation}
\end{proposition}

The gradient of $\mathcal{L}_\text{wBCE}$ is the weighted version of BCE. For all three loss functions, the weights of anchors involve learnable similarity element $s_{ij}$, bringing the flexibility of anchor significance and enabling a more compact and data-efficient learning system. The derivation of the equations and propositions above are given in the Appendix~Sec.~\ref{app-sec:loss-gradient}.

\subsection{Low Rank Similarity Mining}

\begin{figure}[t]
  \begin{center}
  \centerline{\includegraphics[width=0.95\columnwidth]{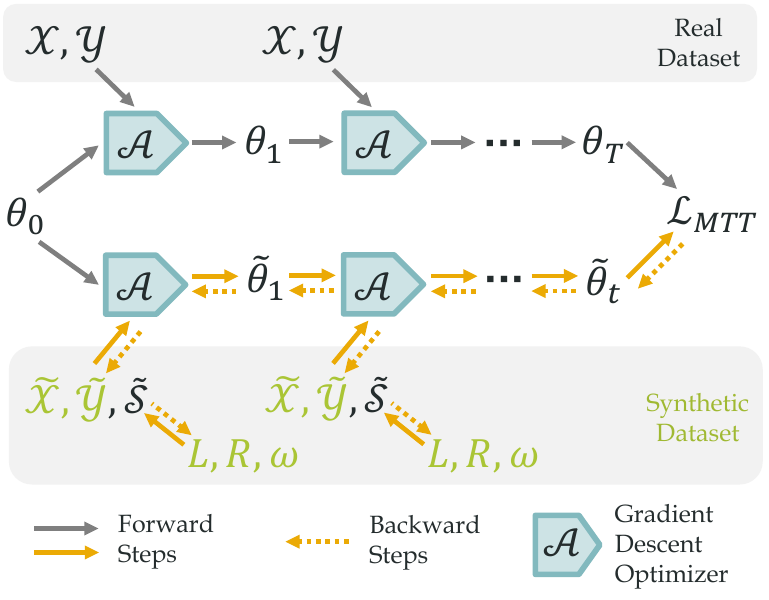}}
  \vspace{-5px}
  \caption{\added{Computation graph of the proposed method \textbf{LoRS}. The green nodes are part of the learnable synthetic dataset.}}
  \label{fig:method}
  \vspace{-10px}
  \end{center}
\end{figure}

Though similarity mining could help the dataset distillation task, the size of extra storage of the similarity matrix increases quadratically and could even exceed the image and text storage when the data size is large. 
The large similarity matrix will also be tricky to optimize and more training iterations are required to train the matrix fully.
Therefore, we exploit the low-rank nature of the similarity matrix to reduce the storage burden.

Ideally, from the perspective of false negative mining, the similarity matrix is instinctively low-rank: if two samples are similar, the two rows or columns in the similarity matrix will be similar according to triangle inequality, which induces a low-rank similarity matrix (Appendix~Sec.~\ref{app-sec:lowrank-nature}).
However, the learned similarity matrix is not and we hope our method could model similarity matrices of all different ranks, including the simplest but full rank identity similarity matrix.
So following~\cite{lora}, we propose to apply low-rank approximation to the \textit{residual similarity matrix}, \ie, we factorize the similarity matrix $\tilde{S}$ into learnable diagonal and low-rank residual matrix:
\begin{equation}
    \tilde{S}=\omega I+\frac{\alpha}{r}L R^\top,
\end{equation}
where the diagonal weight $\omega\in \mathbb{R}^N$ and the low rank components $L, R\in \mathbb{R}^{N\times r}$ are learnable parameters. $r$ is the rank of the residual similarity and $\alpha$ is a weighting factor, which is tuned as hyper-parameters.
Therefore the parameter size of the similarity matrix is $N(2r+1)$, which is linear to the data size. 
In practice, we carefully select a small $r$ and reduce the number of synthetic pairs to maintain the number of learnable parameters of the synthetic data, for a fair comparison to the baseline methods.
The $\omega$ is initialized to ones, the $L$ is initialized randomly and $R$ is initialized to zero, to initialize $S=I$ and avoid the gradient vanishing.
We also empirically analyze the learned similarity matrix to justify the low-rank technique in Appendix~Sec.~\ref{app-sec:eigenvalue} and Fig.~\ref{fig:eigen-analysis} due to the page limit.

Overall, we propose the \textbf{Lo}w-\textbf{R}ank \textbf{S}imilarity Mining (\textbf{LoRS}) technique for the image-text dataset distillation. This approach introduces a new component to the paired multimodal data but can be seamlessly embedded into all multimodal contrastive learning algorithms. 
\added{The overview of the computation graph is also shown in Fig.~\ref{fig:method}. The learnable parameters of the synthetic data are $\tilde{\mathcal{X}},\tilde{\mathcal{Y}}, L, R, \omega$, where $L, R, \omega$ are first composed to the synthetic similarity matrix $\tilde{\mathcal{S}}$, and then used to update the network parameters $\theta$ for the synthetic trajectory. The synthetic and real trajectories are aligned with MTT loss and update the 5 parameters by backpropagation.}
The algorithms are also summarized as Alg.~\ref{alg:lors} taking $\mathcal{L}_{wBCE}$ as an example, and the usage of synthetic data from LoRS is given in Alg.~\ref{alg:eval}.

\begin{algorithm}[t]
    \renewcommand{\algorithmicrequire}{\textbf{Input:}}
    \renewcommand{\algorithmicensure}{\textbf{Output:}}
    \caption{\textbf{Lo}w-\textbf{R}ank \textbf{S}imilarity Mining (\textbf{LoRS})}
    \label{alg:lors}
    \begin{algorithmic}[1]
    \REQUIRE Real data $\mathcal{X}, \mathcal{Y}$
    \ENSURE Synthetic data $\tilde{\mathcal{X}}, \tilde{\mathcal{Y}}$, synthetic similarity matrix $\tilde{S}=\omega I+\frac{\alpha}{r}L R^\top$
    \STATE Initialize $\tilde{\mathcal{X}}, \tilde{\mathcal{Y}}$ by real data, $\omega=\mathbf{1}, L\sim \mathcal{N}, R=0$,
    \REPEAT
        \STATE Sample an initial network parameter $\theta_0$ 
        \STATE Train the network for $T$ steps on $\mathcal{X}, \mathcal{Y}$ to $\theta_T$
        \STATE Train the network for $t$ steps on $\tilde{\mathcal{X}}, \tilde{\mathcal{Y}}, \tilde{S}$ for $t$ steps to $\tilde{\theta}_t$ with $\mathcal{L}_\text{wBCE}$, and keep the computation graph
        \STATE Compute MTT loss $\mathcal{L}_{MTT} = \|\tilde{\theta}_t - \theta_T\|^2  /  \|\theta_0-\theta_T\|^2$
        \STATE $X \gets X - \gamma_X\nabla_X \mathcal{L}$; $Y \gets Y - \gamma_Y\nabla_Y \mathcal{L}$ \\
         $L \gets L - \gamma_s\nabla_L \mathcal{L}$, $R \gets R - \gamma_s\nabla_R \mathcal{L}$ \\
         $\omega \gets \omega - \gamma_s\nabla_\omega \mathcal{L}$ 
    \UNTIL{convergence}
    \end{algorithmic}
\end{algorithm}

\begin{algorithm}[t]
    \renewcommand{\algorithmicrequire}{\textbf{Input:}}
    \renewcommand{\algorithmicensure}{\textbf{Output:}}
    \caption{Train a network with LoRS synthetic data}
    \label{alg:eval}
\begin{algorithmic}[1]
    \REQUIRE Synthetic data $\tilde{\mathcal{X}}, \tilde{\mathcal{Y}}$, synthetic similarity matrix $\tilde{S}$
    \ENSURE A trained network $\theta$
    \STATE Random initialize network $\theta$
    \REPEAT
        \STATE Sample a batch $\{\tilde{x},\tilde{y}\}_{m}$ and their similarity $\{\tilde{s}\}_{m\times m}$
        \STATE Compute $\mathcal{L}_\text{wBCE}$ loss $\mathcal{L}_\text{wBCE}(\{\tilde{x},\tilde{y}\}_{m}, \{\tilde{s}_{ij}\})$
        \STATE $\theta \gets \theta - \gamma\nabla_\theta \mathcal{L}_\text{wBCE}$
    \UNTIL{convergence}
    \end{algorithmic}
\end{algorithm}

\begin{table*}[ht]
\caption{Results on Flickr30k~\cite{Flickr}. The model trained on full dataset performs: IR@1=27.3, IR@5=57.1, IR@10=69.7; TR@1=33.9, TR@5=65.1, TR@10=75.2.}
\label{tab:result-main-flickr}
\begin{center}
\begin{small}
\begin{sc}
\resizebox{0.92\textwidth}{!}{%
    \begin{tabular}{lcl|cccc|cc|c}
    \toprule
    \multirow{2}{*}{Pairs} & \multirow{2}{*}{Ratio} & \multirow{2}{*}{Metric} & \multicolumn{4}{c|}{Coreset Selection} & \multicolumn{3}{c}{Dataset Distillation} \\
    \cmidrule(lr){4-10}
      &  &  & Rand & Herd & K-Cent & Forget & MTT-VL & TESLA$_\text{wBCE}$ & LoRS$_\text{wBCE}$ \\
    \midrule
    \multirow{6}{*}{\shortstack{100\\ (99$\rightarrow$LoRS)}} & \multirow{6}{*}{0.3\%}
      & IR@1  & 1.0  & 0.7 & 0.7 & 0.7 &  4.7$\pm$0.2 &  0.5$\pm$0.2  &  \textbf{ 8.3$\pm$0.2} \\
    & & IR@5  & 4.0  & 2.8 & 3.1 & 2.4 & 15.7$\pm$0.5 &  2.3$\pm$0.2  &  \textbf{24.1$\pm$0.2} \\
    & & IR@10 & 6.5  & 5.3 & 6.1 & 5.6 & 24.6$\pm$1.0 &  4.7$\pm$0.4  &  \textbf{35.1$\pm$0.3} \\
    & & TR@1  & 1.3  & 1.1 & 0.6 & 1.2 &  9.9$\pm$0.3 &  5.5$\pm$0.5  &  \textbf{11.8$\pm$0.2} \\
    & & TR@5  & 5.9  & 4.7 & 5.0 & 4.2 & 28.3$\pm$0.5 & 19.5$\pm$0.9  &  \textbf{35.8$\pm$0.6} \\
    & & TR@10 & 10.1 & 7.9 & 7.6 & 9.7 & 39.1$\pm$0.7 & 28.9$\pm$1.0  &  \textbf{49.2$\pm$0.5} \\
    \midrule
    \multirow{6}{*}{\shortstack{200\\ (199 $\rightarrow$ LoRS)}} & \multirow{6}{*}{0.7\%}
      & IR@1  & 1.1  & 1.5  & 1.5  & 1.2  &  4.6$\pm$0.9 &  0.2$\pm$0.1  &  \textbf{ 8.6$\pm$0.3} \\
    & & IR@5  & 4.8  & 5.5  & 5.4  & 3.1  & 16.0$\pm$1.6 &  1.3$\pm$0.2  &  \textbf{25.3$\pm$0.2} \\
    & & IR@10 & 9.2  & 9.3  & 9.9  & 8.4  & 25.5$\pm$2.6 &  2.5$\pm$0.2  &  \textbf{36.6$\pm$0.3} \\
    & & TR@1  & 2.1  & 2.3  & 2.2  & 1.5  & 10.2$\pm$0.8 &  2.8$\pm$0.5  &  \textbf{14.5$\pm$0.5} \\
    & & TR@5  & 8.7  & 8.4  & 8.2  & 8.4  & 28.7$\pm$1.0 & 10.4$\pm$1.5  &  \textbf{38.7$\pm$0.5} \\
    & & TR@10 & 13.2 & 14.4 & 13.5 & 10.2 & 41.9$\pm$1.9 & 17.4$\pm$1.6  &  \textbf{53.4$\pm$0.5} \\
    \midrule
    \multirow{6}{*}{\shortstack{500\\ (499 $\rightarrow$ LoRS)}} & \multirow{6}{*}{1.7\%}
      & IR@1  & 2.4  &3.0  & 3.5  & 1.8  &  6.6$\pm$0.3 &  1.1$\pm$0.2  & \textbf{10.0$\pm$0.2} \\ 
    & & IR@5  & 10.5 &10.0 & 10.4 & 9.0  & 20.2$\pm$1.2 &  7.3$\pm$0.4  & \textbf{28.9$\pm$0.7} \\ 
    & & IR@10 & 17.4 &17.0 & 17.3 & 15.9 & 30.0$\pm$2.1 & 12.6$\pm$0.5  & \textbf{41.6$\pm$0.6} \\ 
    & & TR@1  & 5.2  &5.1  & 4.9  & 3.6  & 13.3$\pm$0.6 &  5.1$\pm$0.2  & \textbf{15.5$\pm$0.7} \\ 
    & & TR@5  & 18.3 &16.4 & 16.4 & 12.3 & 32.8$\pm$1.8 & 15.3$\pm$0.5  & \textbf{39.8$\pm$0.4} \\ 
    & & TR@10 & 25.7 &24.3 & 23.3 & 19.3 & 46.8$\pm$0.8 & 23.8$\pm$0.3  & \textbf{53.7$\pm$0.3} \\ 
    \bottomrule
    \end{tabular}%
}
\end{sc}
\end{small}
\end{center}
\vspace{-8px}
\end{table*}

\begin{table*}[ht]
\caption{Results on COCO~\cite{COCO}. The model trained on full dataset performs: IR@1=16.9, IR@5=41.9, IR@10=55.9; TR@1=19.6, TR@5=45.6, TR@10=59.5.}
\label{tab:result-main-coco}
\begin{center}
\begin{small}
\begin{sc}
\resizebox{0.92\textwidth}{!}{%
    \begin{tabular}{lcl|cccc|cc|c}
    \toprule
    \multirow{2}{*}{Pairs} & \multirow{2}{*}{Ratio} & \multirow{2}{*}{Metric} & \multicolumn{4}{c|}{Coreset Selection} & \multicolumn{3}{c}{Dataset Distillation} \\
    \cmidrule(lr){4-10}
      &  &  & Rand & Herd & K-Cent & Forget & MTT-VL & TESLA$_\text{wBCE}$ & LoRS$_\text{wBCE}$ \\
    \midrule
    \multirow{6}{*}{\shortstack{100\\ (99$\rightarrow$LoRS)}} & \multirow{6}{*}{0.8\textperthousand}
      & IR@1  & 0.3 & 0.5 & 0.4 & 0.3 &  1.3$\pm$0.1 & 0.3$\pm$0.2  & \bf  1.8$\pm$0.1  \\ 
    & & IR@5  & 1.3 & 1.4 & 1.4 & 1.5 &  5.4$\pm$0.3 & 1.0$\pm$0.4  & \bf  7.1$\pm$0.2  \\ 
    & & IR@10 & 2.7 & 3.5 & 2.5 & 2.5 &  9.5$\pm$0.5 & 1.8$\pm$0.5  & \bf 12.2$\pm$0.2  \\ 
    & & TR@1  & 0.8 & 0.8 & 1.4 & 0.7 &  2.5$\pm$0.3 & 2.0$\pm$0.2  & \bf  3.3$\pm$0.2  \\ 
    & & TR@5  & 3.0 & 2.1 & 3.7 & 2.6 & 10.0$\pm$0.5 & 7.7$\pm$0.5  & \bf 12.2$\pm$0.3  \\ 
    & & TR@10 & 5.0 & 4.9 & 5.5 & 4.8 & 15.7$\pm$0.4 &13.5$\pm$0.3  & \bf 19.6$\pm$0.3  \\ 
    \midrule
    \multirow{6}{*}{\shortstack{200\\ (199 $\rightarrow$ LoRS)}} & \multirow{6}{*}{1.7\textperthousand}
      & IR@1  & 0.6 & 0.9 & 0.7 & 0.6 &  1.7$\pm$0.1 &  0.1$\pm$0.1 &  \bf  2.4$\pm$0.1  \\
    & & IR@5  & 2.3 & 2.4 & 2.1 & 2.8 &  6.5$\pm$0.4 &  0.2$\pm$0.1 &  \bf  9.3$\pm$0.2  \\
    & & IR@10 & 4.4 & 4.1 & 5.8 & 4.9 & 12.3$\pm$0.8 &  0.5$\pm$0.1 &  \bf 15.5$\pm$0.2  \\
    & & TR@1  & 1.0 & 1.0 & 1.2 & 1.1 &  3.3$\pm$0.2 &  0.7$\pm$0.2 &  \bf  4.3$\pm$0.1  \\
    & & TR@5  & 4.0 & 3.6 & 3.8 & 3.5 & 11.9$\pm$0.6 &  3.1$\pm$0.5 &  \bf 14.2$\pm$0.3  \\
    & & TR@10 & 7.2 & 7.7 & 7.5 & 7.0 & 19.4$\pm$1.2 &  5.3$\pm$0.8 &  \bf 22.6$\pm$0.2  \\
    \midrule
    \multirow{6}{*}{\shortstack{500\\ (499 $\rightarrow$ LoRS)}} & \multirow{6}{*}{4.4\textperthousand}
      & IR@1  & 1.1 & 1.7 & 1.1  & 0.8 &  2.5$\pm$0.5  &  0.8$\pm$0.2 & \bf 2.8$\pm$0.2  \\
    & & IR@5  & 5.0 & 5.3 & 6.3  & 5.8 &  8.9$\pm$0.7  &  3.6$\pm$0.6 & \bf 9.9$\pm$0.5  \\
    & & IR@10 & 8.7 & 9.9 & 10.5 & 8.2 & 15.8$\pm$1.5  &  6.7$\pm$0.9 & \bf16.5$\pm$0.7  \\
    & & TR@1  & 1.9 & 1.9 & 2.5  & 2.1 &  5.0$\pm$0.4  &  1.7$\pm$0.4 & \bf 5.3$\pm$0.5  \\
    & & TR@5  & 7.5 & 7.8 & 8.7  & 8.2 & 17.2$\pm$1.3  &  5.9$\pm$0.8 & \bf 18.3$\pm$1.5 \\
    & & TR@10 & 12.5& 13.7& 14.3 & 13.0& 26.0$\pm$1.9  & 10.2$\pm$1.0 & \bf 27.9$\pm$1.4 \\
    \bottomrule
    \end{tabular}%
}
\end{sc}
\end{small}
\end{center}
\vspace{-10px}
\end{table*}

\begin{table*}[t]
\caption{\added{Cross architecture generalization. The data is synthesized with NFNET+BERT and evaluated on various architectures.}}
\label{tab:cross-arch}
\begin{center}
\begin{small}
\begin{sc}
\resizebox{0.93\linewidth}{!}{%
    \begin{tabular}{lll|cccccc}
    \toprule
     \multirow{2}{*}{Pairs} & \multirow{2}{*}{Method} & \multirow{2}{*}{Evaluate Model} & \multicolumn{6}{c}{Flickr} \\ 
    & & & IR@1 & IR@5 & IR@10& TR@1 & TR@5 & TR@10  \\
    \midrule
    \multirow{3}{*}{499} & \multirow{3}{*}{TESLA$_\text{NCE}$}
     & NFNet+BERT   & 8.2$\pm$0.1 & 23.9$\pm$0.4 & 34.7$\pm$0.4 &13.0$\pm$0.5 & 34.5$\pm$0.5 & 49.4$\pm$0.5 \\
    && ResNet+BERT  & 3.0$\pm$0.2 & 10.8$\pm$0.5 & 17.0$\pm$0.8 & 6.0$\pm$0.9 & 18.8$\pm$0.7 & 27.7$\pm$1.2  \\
    && RegNet+BERT  & 3.2$\pm$0.8 & 11.1$\pm$1.8 & 17.5$\pm$1.3 & 5.8$\pm$0.1 & 18.6$\pm$0.6 & 28.1$\pm$1.0 \\
    \midrule
    \multirow{3}{*}{499} & \multirow{3}{*}{LoRS$_\text{wBCE}$}
     & NFNet+BERT   & 10.0$\pm$0.2 & 28.9$\pm$0.7 & 41.6$\pm$0.6 & 15.5$\pm$0.7 & 39.8$\pm$0.4 & 53.7$\pm$0.3 \\
    && ResNet+BERT  &  3.3$\pm$0.2 & 12.7$\pm$0.3 & 20.4$\pm$0.2 &  6.8$\pm$0.2 & 19.6$\pm$1.3 & 31.1$\pm$0.3 \\
    && RegNet+BERT  &  3.5$\pm$0.1 & 12.6$\pm$0.3 & 21.1$\pm$0.4 &  6.8$\pm$0.3 & 20.8$\pm$0.3 & 30.2$\pm$0.3 \\
    \bottomrule
    \end{tabular}%
}
\end{sc}
\end{small}
\end{center}
\end{table*}

\section{Experiments}

\subsection{Dataset and Metrics}
\label{sec:data-metric}

We evaluate our method on Flickr30k~\cite{Flickr} and COCO~\cite{COCO} following the strong baseline~\cite{vl-distill} which exploits the MTT~\cite{MTT} algorithm.
Flickr30k and COCO are image captioning datasets with 31K and 123K images respectively, and each image is paired with five captions. 
The model performance is commonly measured by the recall of top-K retrieval (\textbf{R@K}): given a query from one modality, we retrieve the closest k matches from the other modality and measure the correctness. 
We denote the text-to-image retrieval as \textbf{IR@K} and image-to-text retrieval as \textbf{TR@K}.

\subsection{Baselines and Proposed Method}

We compare various baselines, involving:

(1) \textbf{Coreset Selection}: Random (random select a data subset), Herd~\cite{c-herd}, K-center~\cite{c-kcenter} and Forgetting~\cite{c-forget}.

(2) \textbf{Dataset distillation}: 
MTT-VL~\cite{vl-distill} adapts MTT~\cite{MTT} to image-text pairs (or namely MTT$_\text{NCE}$).
TESLA~\cite{tesla} is an efficient implementation of MTT, so we also adapt TESLA to multimodal data with wBCE loss (TESLA$_\text{wBCE}$).

In comparison, we apply our LoRS technique to the TESLA~\cite{tesla} algorithm with weighted BCE loss (LoRS$_\text{wBCE}$).

\begin{figure}[t]
  \begin{center}
  \centerline{\includegraphics[width=0.98\columnwidth]{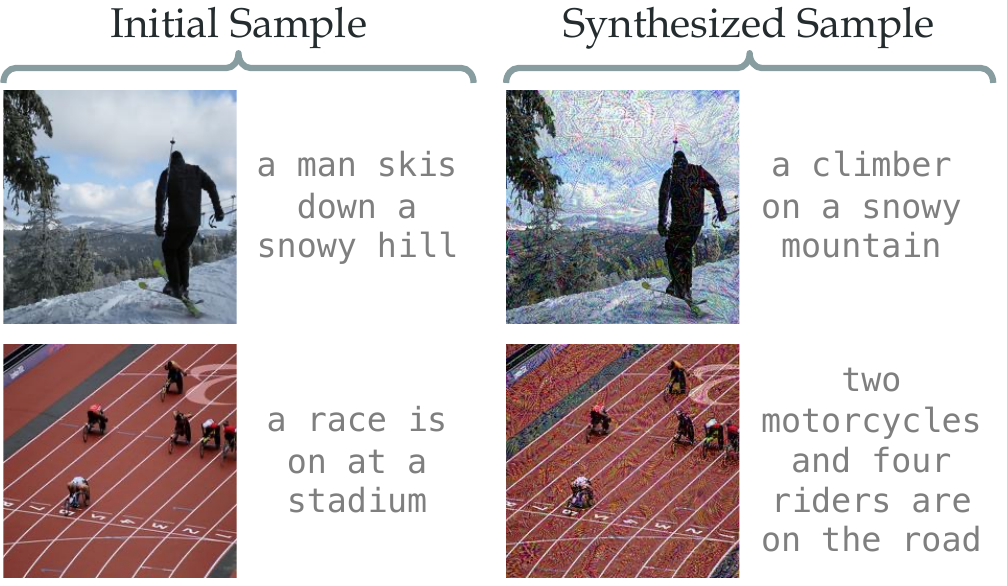}}
  \vspace{-3px}
  \caption{Examples of initialization and synthetic image-text pairs.}
  \label{fig:distilled-data}
  \vspace{-15px}
  \end{center}
\end{figure}

\subsection{Implementation Details}

Following the setting of MTT strong baseline~\cite{vl-distill}, we adopt ImageNet~\cite{imagenet} pretrained NormalizerFree ResNet (NFNet)~\cite{brock2021high} as image encoder and pretrained BERT-base~\cite{devlin2018bert} as text encoder. A linear layer is attached after the text encoder. At both the distillation and training stages, the pretrained weights are loaded and the text network is frozen for efficiency. We directly synthesize the text embedding rather than captions.
We use TESLA~\cite{tesla} as the base distillation algorithm without label learning. We train the network on the full real dataset for 10 epochs 20 times as the expert trajectories.
The experiments are conducted on one RTX 4090 GPU, revealing the efficiency of the method.

In the distillation stage, the images are resized to 224$\times$224 and the text embeddings are 768-d. the synthetic data is learned with SGD and momentum 0.5. The image and text are initialized with random real samples. 
The rest hyper-parameters including learning rate and LoRS parameters vary among different datasets and synthetic data sizes, which are listed in Appendix~Sec.~\ref{app-sec:param} due to page limit.
Particularly, for a fair comparison, we reduce the number of synthetic pairs for LoRS to maintain the synthetic parameters, \eg for experiments with pairs=500, we reduce the number of pairs to 499 for LoRS to save $3\times224^2+768=151K$ parameters, which supports a maximum rank of $r=150$. In practice, we use a smaller rank for efficiency, usually $<50$.

\subsection{Results}

The results on Flickr30k and COCO are shown in Tab.~\ref{tab:result-main-flickr} and \ref{tab:result-main-coco}.
Compared to baselines, LoRS enhances the image-text distillation algorithms and could bring up to $\sim 50\%$ relative improvement. It is interesting that on Flickr30k, LoRS$_\text{wBCE}$ with 100 pairs significantly outperforms the MTT baseline with 500 pairs, showing the larger compression ratio of the similarity mining technique.
It is worth noting that though LoRS changes the data structure, it only brings negligibly $0.3\%$ memory and $0.8\%$ training time overhead. 
\added{For more analysis on the efficiency please refer to Appendix Sec.~\ref{sec:efficiency}.
And the algorithm performance is more significant on Flickr30k since COCO is 3$\times$ larger than Flickr30k and has more complex data relationships.
}

\subsection{Cross Architecture Generalization}

\added{
Following MTT~\cite{MTT}, we conduct a cross-architecture evaluation to study the generalization of the synthetic data. We distill the data with NFNet+BERT and evaluate it with other networks including RegNet~\cite{regnet} and ResNet50~\cite{resnet}. 
It is not necessary to validate the generalization of the text network as we freeze the text encoder.
Results in Tab.~\ref{tab:cross-arch} show that our distilled data could generalize across networks (significantly surpasses the coreset selection methods in Tab.~\ref{tab:result-main-flickr}), and also outperform the baseline model. Note that the performance drop is also partly due to the performance of the architectures themselves (\eg ResNet or RegNet trained on full data achieves about IR@1=28\% and TR@1=22\%, while NFNet achieves about IR@1=33\% and TR@1=27\%).
}

\begin{table*}[t]
\caption{Various ablation studies with 200 pairs on Flickr30k~\cite{Flickr} and COCO~\cite{COCO}.}
\vspace{-2mm}
\label{tab:ablation}
\begin{center}
\begin{small}
\begin{sc}
\resizebox{\textwidth}{!}{%
    \begin{tabular}{ll|cccccc|cccccc}
    \toprule
    \multirow{2}{*}{No.} & \multirow{2}{*}{Model} & \multicolumn{6}{c|}{Flickr} & \multicolumn{6}{c}{COCO} \\
    & & IR@1 & IR@5 & IR@10& TR@1 & TR@5 & TR@10 & IR@1 & IR@5 & IR@10& TR@1 & TR@5 & TR@10 \\
    \midrule
    (1) & Similarity Mining+BCE &  0.6 &  3.8 &  6.6 &  1.2 &  4.4 &  7.1 & 0.0 & 0.1 & 0.2 & 0.0 & 0.1 & 0.3 \\
    (2) & Similarity Mining+eNCE&  4.1 & 13.4 & 20.8 &  5.5 & 17.3 & 26.3 & 1.3 & 5.1 & 9.0 & 1.9 & 7.1 & 11.7 \\
    (3) & Similarity Mining+wBCE&  7.4 & 23.9 & 35.3 & \bf15.1 & 37.7 & 51.4 & 2.2 & 8.1 & 14.0 & 4.1 & 13.7 & 21.7 \\
    \midrule
    (4) & LoRS$_\text{BCE}$     &  0.1 &  0.6 &  1.0 &  0.2 &  0.7 &  1.2 & 0.0 & 0.1 & 0.2 & 0.1 & 0.7 & 1.5 \\
    (5) & LoRS$_\text{eNCE}$    &  8.2 & \bf25.7 & \bf37.5 & 13.0 & 36.3 & 51.0 & 2.0 & 7.7 & 13.0 & 3.9 & 13.9 & 22.2 \\
    (6) & LoRS$_\text{wBCE}$    &  \bf8.6 & 25.3 & 36.6 & 14.5 & \bf38.7 & \bf53.4 & \bf2.4 & \bf9.3 & \bf15.5 & \bf4.3 & \bf14.2 & \bf22.6 \\
    \midrule
    (7) & LoRS$_\text{wBCE}$, $r=1$  & 7.2 & 22.6 & 33.8 & 13.1 & 37.6 & 51.7 & 1.8 & 6.8 & 11.8 & 3.5 & 11.9 & 19.5 \\
    (8) & LoRS$_\text{wBCE}$, $r=5$  & \bf8.6 & 25.3 & 36.6 & 14.5 & \bf38.7 & \bf53.4 & 1.6 & 6.8 & 12.1 & 2.9 & 10.6 & 17.7 \\
    (9) & LoRS$_\text{wBCE}$, $r=10$ & 7.8 & 24.1 & 36.2 & 14.4 & 37.9 & 51.9 & 2.3 & 8.4 & 14.3 & 4.1 & 13.5 & 21.5 \\
    (10)& LoRS$_\text{wBCE}$, $r=20$ & 7.9 & 24.0 & 35.7 & 13.7 & 37.8 & 50.2 & \bf2.4 & \bf9.3 & \bf15.5 & \bf4.3 & \bf14.2 & \bf22.6 \\
    (11)& LoRS$_\text{wBCE}$, $r=40$ & 7.5 & 23.6 & 35.2 & 14.6 & 38.2 & 52.0 & 2.3 & 8.5 & 14.5 & 3.9 & 13.9 & 22.1 \\
    \midrule
    (12)& LoRS$_\text{wBCE}$ w/o $L,~R$         &  7.4 & 23.4 & 36.0 & 13.9 & 38.1 & 52.2 & 2.3 & 8.3 & 14.0 & 3.8 & 13.6 & 21.9 \\
    (13)& LoRS$_\text{wBCE}$ w/o $L,~R,~\omega$ &  0.2 &  1.3 &  2.5 &  2.8 & 10.4 & 17.4 & 0.1 &  0.2 & 0.5 & 0.7 &  3.1 & 5.3  \\
    \midrule
    (14)& LoRS$_\text{wBCE}$, fix image          & 4.9 & 16.3 & 24.9 & 9.8 & 29.4 & 41.1 &  1.4 & 5.3 & 9.3 & 1.9 & 6.6 & 11.7 \\
    (15)& LoRS$_\text{wBCE}$, fix text           & 4.3 & 15.5 & 25.2 & 9.2 & 24.7 & 36.3 &  0.6 & 2.9 & 5.3 & 1.5 & 5.8 & 9.9 \\
    (16)& LoRS$_\text{wBCE}$, fix image+text & 1.9 &  8.2 & 13.9 & 4.3 & 12.8 & 19.9 &  0.6 & 2.5 & 4.7 & 1.2 & 5.2 & 8.8 \\
    
    \midrule
    (17)& CLIP Similarity & 7.4 & 22.8 & 33.8 & 10.9 & 32.4 & 44.0 & 1.8 & 7.4 & 12.7 & 2.8 & 10.6 & 17.6 \\ 
    \bottomrule
    \end{tabular}%
}
\end{sc}
\end{small}
\vspace{-8px}
\end{center}
\end{table*}

\subsection{Ablation Study}

Tab.~\ref{tab:ablation} shows the results of the ablation study.

\noindent{\bf Learn full similarity matrix (No.~1-3).} We implemented similarity mining with full learnable similarity matrix ($\tilde{N}\times \tilde{N}$ parameters, without the low-rank technique). The full similarity mining shows comparable performance with LoRS, indicating the feasibility of the low-rank approximation of the similarity matrix.\\
\noindent{\bf Losses (No.~4-6).} \added{Among the losses, $\mathcal{L}_\text{wBCE}$ slightly surpasses $\mathcal{L}_\text{eNCE}$, while significantly outperforms vanilla $\mathcal{L}_\text{BCE}$, mainly due to their balancedness. Along with the comparison of $\mathcal{L}_\text{NCE}$ in Tab.~\ref{tab:result-main-flickr} and \ref{tab:result-main-coco}, we suggest choosing between $\mathcal{L}_\text{wBCE}$ and $\mathcal{L}_\text{eNCE}$ for LoRS.} \\
\noindent{\bf Rank $r$ (No.~7-11).} As long as $r$ is not too small, it slightly affects the performance and $r=20$ is sufficient here.\\
\noindent{\bf Components in low-rank factorization (No.~12-13).} Removing the low-rank components $L, R$ reduces the performance but still surpasses the one with an identity matrix (No.~13). \\
\noindent{\bf Fix image or text (No.~14-16).} Freezing the image or text during distillation could greatly reduce the data performance and experiments show that learning text is more critical for the distillation. It is surprising that on Flickr30k, the experiment that only learns the similarity matrix (No.~16) could perform above random model. \\
\noindent{\bf Similarity from pretrained CLIP (No.~17).} Instead of \textit{learning} a similarity matrix, we directly compute the similarity matrix with a pretrained CLIP. However, the computed similarity matrix does not fit the distilled image and text, resulting in poor retrieval performance. This phenomenon is in line with the common conclusion in dataset distillation: the data that is suitable for network training may not be natural to humans.

\begin{figure}[t]
    \centering
    \begin{subfigure}{0.35\columnwidth}
        \centering
        \includegraphics[width=\columnwidth]{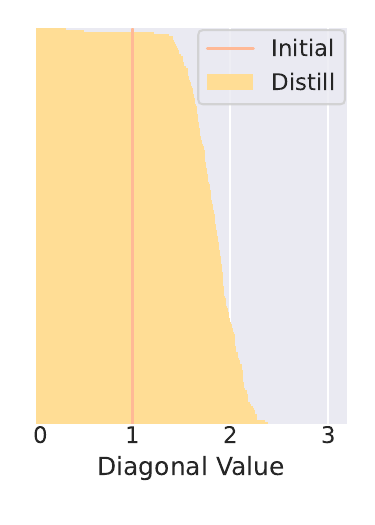}
        \caption{Diagonal $\omega$}
    \end{subfigure}
    \hfill
    \begin{subfigure}{0.6\columnwidth}
        \centering
        \includegraphics[width=\columnwidth]{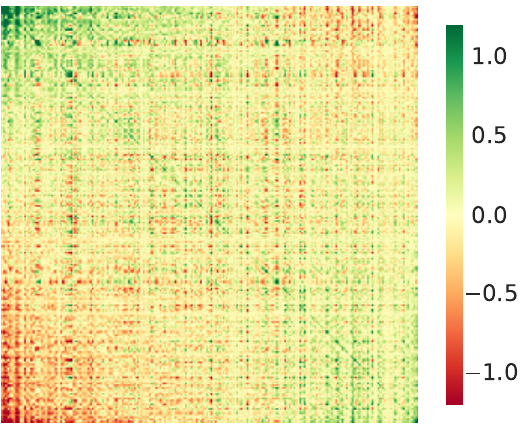}
        \caption{Residual Similarity $LR^\top$}
    \end{subfigure}
    \caption{Synthesized similarity matrix $\tilde{S}$.}
    \label{fig:visualize-mat}
\end{figure}

\begin{figure}[t]
    \centering
  \begin{center}
  \centerline{\includegraphics[width=\columnwidth]{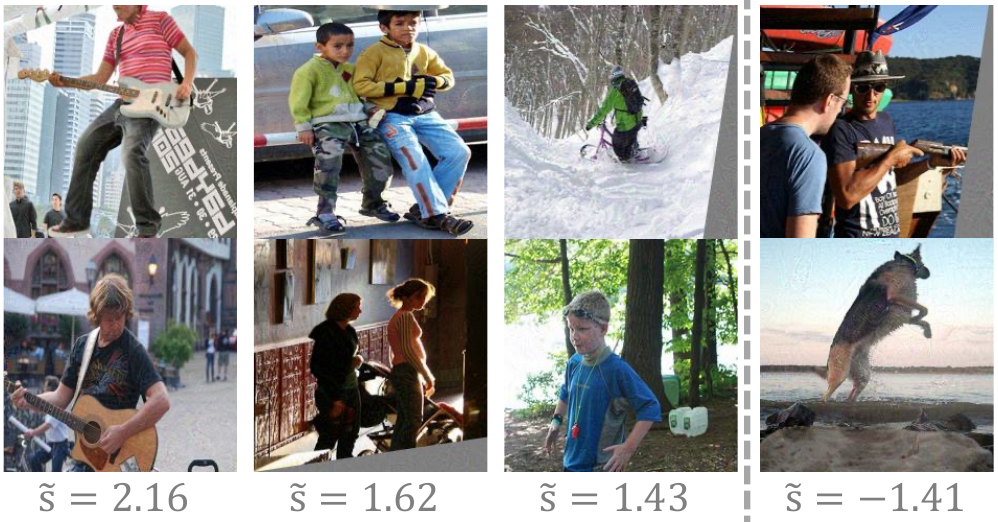}}
  \caption{Sample pairs with different synthetic similarity $\tilde{s}$.}
  \label{fig:visualize-falseneg}
  \end{center}
  \vspace{-3mm}
\end{figure}

\subsection{Visualization}

We visualize the image, text, and similarity matrix of 200 synthetic pairs for Flickr30k to present the distilled data.

\noindent{\bf Synthetic image and text.}
Fig.~\ref{fig:distilled-data} shows the image and text before (initial) and after distillation. The images get DeepDream-style~\cite{zeiler2014visualizing}, common in dataset distillation. The texts are retrieved by the closest caption in the train set to the distilled embeddings following~\cite{vl-distill}.
Appendix~Sec.~\ref{app-sec:visualize} gives more examples.

\paragraph{Learned similarity matrix}
For clarity, we separately show the diagonal and residual matrix in Fig.~\ref{fig:visualize-mat}. Our method tends to learn large diagonal values since they are positive pairs. LoRS could also find the false negatives by learning certain similarity scores.
We visualize some sample pairs with different synthetic similarities in Fig.~\ref{fig:visualize-falseneg}. The samples that LoRS assigns large similarity values are also similar from a human perspective (left three pairs in the figure, with a similar person, background, etc.), which a regular CLIP model will erroneously regard as negative pairs.

\section{Conclusions}

In this work, we introduce Low-Rank Similarity Mining (LoRS) as an efficient solution for image-text dataset distillation. LoRS concurrently distills a ground truth similarity matrix with image-text pairs, leveraging low-rank factorization for efficiency and scalability. Our approach demonstrates a substantial improvement over existing algorithms. We advocate for the adoption of LoRS as a foundational synthetic data setup for image-text dataset distillation.

\section*{Acknowledgments}
This work is supported in part by the
National Natural Science Foundation of China under Grants No.62306175, No.62302296, 
National Key Research and Development Project of China (No.2022ZD0160102, No.2021ZD0110704).

\section*{Impact Statement}

This paper presents work whose goal is to advance the field of 
Machine Learning, especially multimodal learning. There are many potential societal consequences 
of our work, none which we feel must be specifically highlighted here.

\bibliography{main}
\bibliographystyle{icml2024}

\newpage
\appendix
\onecolumn

\section{Dataset Variance}
\label{sec:variance}

\added{
We compare the variance of image-text (multimodal) datasets and classification datasets to study why some dataset distillation algorithms fail on image-text datasets.
We choose three multimodal datasets (Flickr30k, Visual Genome, and COCO-Caption) and seven image classification datasets (DTD, Caltech-101, MIT-indoor-67, CIFAR-10/100, iNaturalist-18 and ImageNet-1K).
The images are forwarded to CLIP encoders with ResNet or ViT architecture, pretrained on YFCC100M~\cite{yfcc100m} or LAION-5B~\cite{laion-5b}.
Note that we compute the intra-class variance for classification datasets.
The full results are shown in Tab.~\ref{tab:variance} and plotted in Fig.~\ref{fig:data-variance}. The variance of multimodal datasets is significantly larger than classification datasets.
}

\begin{table*}[ht]
\caption{\added{Comparison of dataset variance. CLS=classification dataset; MM=multimodal dataset.}}
\label{tab:variance}
\begin{center}
\begin{small}
\begin{sc}
\resizebox{0.95\textwidth}{!}{%
    \begin{tabular}{l|cr|ccc}
    \toprule
    Dataset & Type & \shortstack{Data\\Scale} & \shortstack{YFCC100M\\ViT-B/32 ($\times 10^{-2}$)} & \shortstack{YFCC100M\\ResNet50 ($\times 10^{-4}$)} & \shortstack{LAION-5B\\ViT-B/32 ($\times 10^{-2}$)} \\
    \midrule
    Flickr30k~\cite{Flickr}  & MM  & 29.0 K   & 9.082  & 15.635 & 12.744 \\
    Visual Genome~\cite{vg}  & MM  & 108.2 K  & 9.543  & 17.832 & 13.578 \\
    COCO-Caption~\cite{COCO} & MM  & 113.3 K  & 9.588  & 17.656 & 13.652 \\
    \midrule
    DTD~\cite{dtd}              & CLS & 5.6 K    & 5.657  &  7.289 &  9.052 \\
    Caltech101~\cite{caltech101}& CLS & 9.1 K    & 4.411  &  7.256 &  6.833 \\
    MIT-indoor~\cite{MIT-indoor}& CLS & 15.6 K   & 5.043  &  8.940 &  7.918 \\
    CIFAR-10~\cite{cifar}       & CLS & 50.0 K   & 3.823  &  4.991 &  7.759 \\
    CIFAR-100~\cite{cifar}      & CLS & 50.0 K   & 3.750  &  4.663 &  7.498 \\
    iNaturalist-18~\cite{inaturalist}& CLS & 461.9 K  & 3.316  &  6.873 &  5.415 \\
    ImageNet-1K~\cite{imagenet} & CLS & 1281.2 K & 5.079  &  6.880 &  7.937 \\
    \bottomrule
    \end{tabular}%
}
\end{sc}
\end{small}
\end{center}
\end{table*}

\section{Derivation of Loss Gradients}
\label{app-sec:loss-gradient}

\subsection{Proposition \ref{prop:NCE} and \ref{prop:eNCE}}

Since we assume representations $u_i$ and $v_j$ are normalized, the cosine similarity $\tilde{s}_{ij}=u_i^\top v_j$. The $\mathcal{L}_\text{eNCE}$:
\begin{equation}
\begin{aligned}
\mathcal{L}_\text{eNCE}(\mathcal{B}, S) 
= & -\frac1m\sum_{i=1}^m\sum_{j=1}^ms_{ij}\left[
\log P^V_{ij}+\log P^{T}_{ij}
\right] \\
= &-\frac1m\sum_{i=1}^m\sum_{j=1}^ms_{ij}\left[
\log\frac{\exp(u_i^\top v_j/\tau)}{\sum_{k=1}^m\exp(u_i^\top v_k/\tau)}
+
\log\frac{\exp(u_i^\top v_j/\tau)}{\sum_{k=1}^m\exp(u_k^\top v_j/\tau)}
\right] \\
= &\frac1m\sum_{i=1}^m\sum_{j=1}^ms_{ij}\left[
\log{\sum_{k=1}^m\exp(u_i^\top v_k/\tau)}
+
\log{\sum_{k=1}^m\exp(u_k^\top v_j/\tau)}
-\frac2\tau u_i^\top v_j
\right] \\
\triangleq & \frac1m(L_A+L_B-L_C).
\end{aligned}
\end{equation}

Then we derive the gradient of each of the three components $L_A, L_B, L_C$:

\begin{equation}
\begin{aligned}
\frac{\partial L_A}{\partial u_n} 
= & \frac{\partial}{\partial u_n} \sum_{i=1}^m\sum_{j=1}^m s_{ij}\log{\sum_{k=1}^m\exp(u_i^\top v_k/\tau)} 
=  \frac{\partial}{\partial u_n} \sum_{j=1}^m s_{nj}\log{\sum_{k=1}^m\exp(u_n^\top v_k/\tau)} \\
= & \left(\frac{\partial}{\partial u_n} \log{\sum_{k=1}^m\exp(u_n^\top v_k/\tau)} \right)\left(\sum_{j=1}^m s_{nj}\right) \\
= & \left(\frac{\frac{\partial}{\partial u_n} {\sum_{k=1}^m\exp(u_n^\top v_k/\tau)}}{\sum_{k'=1}^m\exp(u_n^\top v_{k'}/\tau)} \right)s_{n\bigcdot}
=  \left(\frac{\sum_{k=1}^m\exp(u_n^\top v_k/\tau)v_k/\tau}{\sum_{k'=1}^m\exp(u_n^\top v_{k'}/\tau)} \right)s_{n\bigcdot} \\
= & \left(\sum_{k=1}^m P_{nk}^V \frac{v_k}\tau \right)s_{n\bigcdot}, \\
\end{aligned}
\end{equation}

\begin{equation}
\begin{aligned}
\frac{\partial L_B}{\partial u_n} 
= & \frac{\partial}{\partial u_n} \sum_{i=1}^m\sum_{j=1}^m s_{ij}\log{\sum_{k=1}^m\exp(u_k^\top v_j/\tau)} 
=  \sum_{i=1}^m\sum_{j=1}^m s_{ij}\left(\frac{\partial}{\partial u_n}\log{\sum_{k=1}^m\exp(u_k^\top v_j/\tau)}\right) \\
= & \sum_{i=1}^m\sum_{j=1}^m s_{ij}\left(\frac{\frac{\partial}{\partial u_n}{\sum_{k=1}^m\exp(u_k^\top v_j/\tau)}}{\sum_{k=1}^m\exp(u_k^\top v_j/\tau)} \right)
= \sum_{i=1}^m\sum_{j=1}^m s_{ij}\left(\frac{\exp(u_n^\top v_j/\tau) v_j/\tau}{\sum_{k=1}^m\exp(u_k^\top v_j/\tau)} \right) \\
= & \sum_{i=1}^m\sum_{j=1}^m s_{ij}\left(P_{nj}^T \frac{v_j}\tau \right) 
= \left(\sum_{i=1}^m s_{ij}\right) \sum_{j=1}^m \left(P_{nj}^T \frac{v_j}\tau \right) 
= s_{\bigcdot j} \sum_{j=1}^m \left(P_{nj}^T \frac{v_j}\tau \right).
\end{aligned}
\end{equation}

\begin{equation}
\begin{aligned}
\frac{\partial L_C}{\partial u_n} 
= & \frac{\partial}{\partial u_n} \sum_{i=1}^m\sum_{j=1}^m\frac{2s_{ij}}\tau u_i^\top v_j
= \frac{\partial}{\partial u_n} \sum_{j=1}^m\frac{2s_{nj}}\tau u_n^\top v_j
= \sum_{j=1}^m\frac{2s_{nj}}\tau v_j.
\end{aligned}
\end{equation}

The overall gradient of $\mathcal{L}_\text{eNCE}$ is:

\begin{equation}
\begin{aligned}
\frac{\partial }{\partial u_n} \mathcal{L}_\text{eNCE}(\mathcal{B}, S)
= & \frac1m(L_A+L_B-L_C) = 
\frac1m\left(\sum_{k=1}^m P_{nk}^V \frac{v_k}\tau \right)s_{n\bigcdot} 
+ \frac1m s_{\bigcdot j} \sum_{j=1}^m \left(P_{nj}^T \frac{v_j}\tau \right) 
- \frac1m\sum_{j=1}^m\frac{2s_{nj}}\tau v_j \\
= & \frac1{m\tau}\sum_{j=1}^m \left( 
P_{nj}^V s_{n\bigcdot} + P_{nj}^T s_{\bigcdot j} -2s_{nj}
\right) v_j.
\end{aligned}
\end{equation}

When the similarity matrix is identity, \ie $S=I$, the gradient degrades to:

\begin{equation}
\begin{aligned}
\frac{\partial }{\partial u_n} \mathcal{L}_\text{NCE}(\mathcal{B}, S)
=\frac1{m\tau}\sum_{j=1}^m \left( 
P_{nj}^V + P_{nj}^T -2s_{nj}
\right) v_j
= \begin{cases}
\frac1{m\tau}\sum_{j=1}^m\left( P_{nj}^V + P_{nj}^T -2 \right) v_j, &~\text{if}~~ n=j \\
\frac1{m\tau}\sum_{j=1}^m\left( P_{nj}^V + P_{nj}^T \right) v_j, &~\text{if}~~ n\neq j \\
\end{cases}
\end{aligned}
\end{equation}

\subsection{Proposition \ref{prop:BCE}}

Similar to InfoNCE loss, the BCE loss 

\begin{equation}
\mathcal{L}_\text{BCE}(\mathcal{B}, S)
= \frac1m\sum_{i=1}^m\sum_{j=1}^m \ell\left(s_{ij}, \sigma(\hat{s}_{ij}/\tau)\right)
= \frac1m\sum_{i=1}^m\sum_{j=1}^m \ell\left(s_{ij}, \sigma(u_i^\top v_j/\tau)\right),
\end{equation}
where $\sigma(x)$ is the sigmoid function and $\ell(y,p)=-y\log(p)-(1-y)\log(1-p)$. We first have:

\begin{equation}
\begin{aligned}
\ell\left(y,\sigma(x)\right) 
& = -y\log\frac1{1+e^{-x}}-(1-y)\log\frac{e^{-x}}{1+e^{-x}} \\
& = y\log(1+e^{-x})+(1-y)x+(1-y)\log(1+e^{-x})=\log(1+e^{-x})+(1-y)x.
\end{aligned}
\end{equation}

Thus:

\begin{equation}
\begin{aligned}
& \frac{\partial\ell\left(y,\sigma(x)\right) }{\partial x}=\frac{-e^{-x}}{1+e^{-x}}+(1-y)=\sigma(x)-y.
\end{aligned}
\end{equation}

Therefore the overall gradient of BCE is:

\begin{equation}
\frac{\partial}{\partial u_n} \mathcal{L}_\text{BCE}(\mathcal{B}, S)
= \frac1m\sum_{j=1}^m \frac{\partial}{\partial u_n} \ell\left(s_{nj}, \sigma(u_n^\top v_j/\tau)\right) 
= \frac1{m\tau}\sum_{j=1}^m (\sigma(\tilde{s}_{nj}/\tau)-s_{nj})v_j. 
\end{equation}

\section{Justification of Low-Rank Technique}

\subsection{Low-Rank Nature of the Ideal Similarity Matrix}
\label{app-sec:lowrank-nature}

Given a image-text dataset on embedding space $\{u_i\}, \{v_i\}$ and a distance metric $d(\bigcdot,\bigcdot)$, the distance matrix is $D=\{d_{ij}\}=\{d(u_i, v_j)\}$.
If two images embeddings $u_i, u_j$ are similar, \ie $d(x_i, x_j)<\epsilon$ where $\epsilon>0$ is a small value, according to the triangle inequality, $\forall k$, $|d_{ik}-d_{jk}|=|d(u_i, v_k)-d(u_j, v_k)|\leq d(u_j, u_i)=\epsilon$.
Hence the $i^\text{th}$ row and $j^\text{th}$ row of the distance matrix are similar.
As the similarity metric is always a function of the distance metric, we conclude that once similar samples exist, there are similar rows or columns in the similarity matrix, which leads to a low-rank similarity matrix.

\begin{figure}[t]
  \begin{center}
  \centerline{\includegraphics[width=0.5\columnwidth]{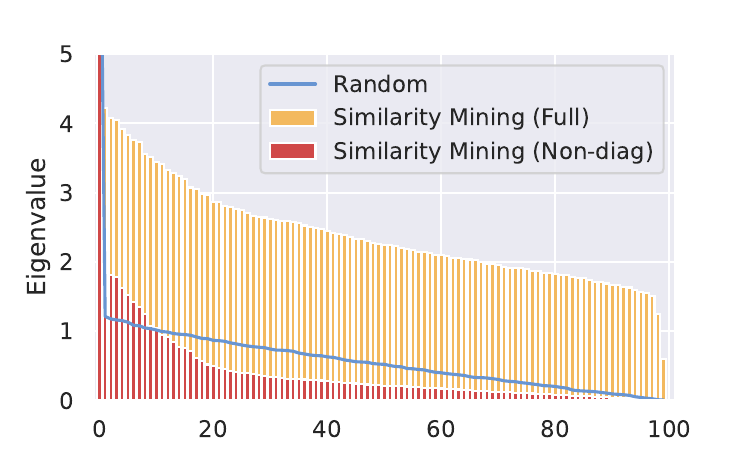}}
  \caption{Eigenvalues of the distilled similarity matrix. The yellow bars are eigenvalues of the full similarity matrix; the red bars are that of the matrix without diagonal; the blue curve is the eigenvalues of a Gaussian noise matrix with the same mean and variance as the matrix of the red bars.}
  \label{fig:eigen-analysis}
  \vspace{-10px}
  \end{center}
\end{figure}

\subsection{Eigenvalue Analysis of Similarity Mining}
\label{app-sec:eigenvalue}

We compute the eigenvalues of the fully learned similarity matrix and the matrix that zero-masked the diagonal.
As shown in Fig.~\ref{fig:eigen-analysis}, the similarity without diagonal has much smaller eigenvalues. Compared to the random Gaussian matrix with the same mean and variance (blue), the residual similarity matrix has a more long-tailed eigenvalue distribution, indicating its low-rank nature.

\section{Extended Ablation Study}

\subsection{Data Filtering}

\added{
Data filtering (\eg~DataComp~\cite{datacomp}) has become a popular and efficient method to reduce the data scale for multimodal and image-text contrastive learning datasets, while it lacks comparison for dataset distillation studies.
To enrich the discussion on image-text dataset distillation, we give an analysis and comparison of coreset selection, data filtering, and dataset distillation, as first summarized in Tab.~\ref{tab:app:comp-data-filter}.
}

\begin{table*}[ht]
\caption{\added{Comparison to of various data size reduction methods.}}
\label{tab:app:comp-data-filter}
\begin{center}
\begin{small}
\begin{sc}
\resizebox{0.65\textwidth}{!}{%
    \begin{tabular}{l|cc}
    \toprule
    Method & \shortstack{From massive noisy data\\To high quality data} & \shortstack{From high quality data\\To smaller data}  \\ 
    \midrule
    Data filtering          & \textbf{fast and good} & fast but worse    \\
    Coreset Selection       & slow          & good              \\
    Dataset Distillation    & slower        & \textbf{best}              \\
    \bottomrule
    \end{tabular}%
}
\end{sc}
\end{small}
\end{center}
\end{table*}

\added{
\textbf{Data filtering \& Coreset selection}. They share similar technical essences but are applied to different scenarios for different objectives of machine learning tasks. Data filtering is finding useful data from noisy large-scale internet/in-the-wild data, usually aiming to improve the data quality. Whilst, coreset selection focuses on smaller but more accurate datasets, only to enhance the training efficiency with a tolerable performance drop.
Most coreset selection algorithms could be directly applied to data filtering, but are inefficient for the large-scale dataset, and most data filtering could be used for coreset selection but may have worse performance.
}

\added{
\textbf{Coreset selection \& Dataset distillation}. The dataset distillation is an ``upgraded'' or learnable version of coreset selection, whose motivations and applications overlap. A fascinating point of dataset distillation is that it is possible to significantly reduce the data size but keep the performance even for a high-quality dataset, which is not possible for data filtering or selection.
}

\added{
\textbf{Experimental analysis}. To extend our comparison, we conduct experiments of data filtering. We adopt the CLIP/LAION score criterion in DataComp with 3 pretrained CLIP models (ResNet50 or ViT-B/32 image encoder; BERT text encoder; YFCC100M~\cite{yfcc100m} or LAION-5B~\cite{laion-5b} pretraining), and the results are presented in Tab.~\ref{tab:app:data-filter-result}. On the high-quality datasets, the data filtering method is comparable to coreset selection, while the dataset distillation method (LoRS) significantly outperforms both.
}

\begin{table*}[t]
\caption{\added{Comparison to data filtering methods. \textbf{Bold}: best method. \underline{Underlined}: second best method.}}
\label{tab:app:data-filter-result}
\begin{center}
\begin{small}
\begin{sc}
\resizebox{0.7\textwidth}{!}{%
    \begin{tabular}{ll|ccc|cc}
    \toprule
    Dataset & Pairs & \shortstack{ResNet-50\\YFCC100M} & \shortstack{ViT-B/32\\YFCC100M} & \shortstack{ViT-B/32\\LAION-5B} & \shortstack{Coreset\\Selection} & LoRS  \\
    \midrule
    \multirow{2}{*}{Flickr30k}  & 100  & 4.42  & 4.40  & 3.41  & \underline{4.80}  & \bf 27.38 \\
                                & 200  & \underline{6.95}  & 6.26  & 6.41  & 6.52  & \bf 29.52 \\
                                & 500  & 12.95 & \underline{13.40} & 12.66 & 13.25 & \bf 31.58 \\
    \midrule
    \multirow{2}{*}{COCO}       & 100  &  1.11 & 1.41  & 1.11  & \underline{2.48}  & \bf 9.37 \\
                                & 200  &  2.20 & 2.63  & 2.00  & \underline{3.52}  & \bf 11.38 \\
                                & 500  &  5.11 & 6.46  & 5.05  & \underline{7.23}  & \bf 13.45 \\
    \bottomrule
    \end{tabular}%

}
\end{sc}
\end{small}
\end{center}
\vspace{-5mm}
\end{table*}

\begin{table*}[t]
\caption{\added{Comparison of various architectures. LoRS consistently outperforms baselines.}}
\label{tab:app:more-arch}
\begin{center}
\begin{small}
\begin{sc}
\resizebox{0.85\textwidth}{!}{%
    \begin{tabular}{ll|cccc|cccc}
    \toprule
    \multirow{2}{*}{Image encoder} & \multirow{2}{*}{Text encoder}  & \multicolumn{4}{c|}{100 pairs} & \multicolumn{4}{c|}{200 pairs} \\
        &   & Rand  & Filter &  MTT-VL  & LoRS & Rand  & Filter  &  MTT-VL  & LoRS  \\
    \midrule
    NFResNet & BERT       & 4.1  & 6.3  & 16.0 & {\bf 20.7} &  5.1 & 7.1  & 17.9 & {\bf 24.2} \\
    ViT      & BERT       & 2.9  & 4.3  & 11.5 & {\bf 16.4} &  2.6 & 4.1  & 12.6 & {\bf 16.7} \\
    NFRegNet & BERT	      & 5.1  & 6.9  & 15.5 & {\bf 22.0} &  4.4 & 6.2  & 18.9 & {\bf 26.2} \\
    NFNet    & DistilBERT & 7.4  & 10.7 & 24.0 & {\bf 26.0} &  6.2 & 10.6 & 24.4 & {\bf 27.8} \\
    NFNet    & CLIP-Text  & 16.0 & 24.5 & 47.6 & {\bf 54.9} & 15.5 & 27.8 & 49.7 & {\bf 60.1} \\
    \bottomrule
    \end{tabular}%
}
\end{sc}
\end{small}
\end{center}
\end{table*}

\subsection{Network Architectures}
\added{
Besides the NFNet+BERT model, we extend our experiments to more image and text encoder structures on Flickr30k~\cite{Flickr} in Tab.~\ref{tab:app:more-arch}. LoRS surpasses baseline methods by a large margin, which shows generalization ability across various architectures.
}

\begin{table*}[ht]
\vspace{-5mm}
\caption{\added{Time comparison. The results are in \textit{second/iteration}.}}
\label{tab:app:time-comp}
\begin{center}
\begin{small}
\begin{sc}
\resizebox{0.55\textwidth}{!}{%
    \begin{tabular}{ll|ccc}
    \toprule
    Dataset     & Method    & 100 pairs     & 200 pairs     & 500 pairs     \\
    \midrule
    \multirow{2}{*}{Flickr30k}  & Baseline	& 6.47$\pm$0.27 & 6.59$\pm$0.48 & 6.43$\pm$0.32  \\
                                & LoRS      & 6.35$\pm$0.13 & 6.52$\pm$0.52 & 6.42$\pm$0.12  \\
    \midrule
    \multirow{2}{*}{COCO}       & Baseline  & 6.27$\pm$0.39 & 6.26$\pm$0.27 & 6.15$\pm$0.20  \\
                                & LoRS      & 6.00$\pm$0.14 & 5.89$\pm$0.07 & 5.92$\pm$0.13  \\
    \bottomrule
    \end{tabular}%
}
\end{sc}
\end{small}
\end{center}
\vspace{-5mm}
\end{table*}

\subsection{Efficiency Analysis}
\label{sec:efficiency}

\added{
    Efficiency counts in dataset distillation. 
    Memory and time cost are also one of the main motivations to leverage the low-rank method, as the storage of the similarity matrix should be considered as part of the synthetic dataset.
    Overall, the LoRS introduces negligible memory and time overhead compared to the large performance gain it brings.
    We analyze the efficiency of LoRS as follows.
    \begin{itemize}
        \item {\bf Time}: we conduct experiments of distillation time comparison in Tab.~\ref{tab:app:time-comp}. Due to the small parameter size of LoRS, it takes \textbf{comparable} time to the baseline method (the difference is smaller than the variance). Note that the synthetic data is optimized in batch so the data size (number of pairs) does not influence the iteration time.
        \item {\bf Memory}: For a fair comparison, we have reduced the number of synthetic pairs only for LoRS in Tab.~\ref{tab:result-main-flickr}~\ref{tab:result-main-coco}. So in these experiments, our method uses less memory storage but achieves significantly higher performance. And with the low-rank method, the memory overhead of LoRS is linear to the data size which is acceptable or even negligible. For example, with $r=50$, the overhead of LoRS is 0.07\% of the total data storage no matter the data scale.
    \end{itemize}   
}

\section{More Visualizations of Synthetic Dataset}
\label{app-sec:visualize}

\begin{figure}[t]
  \begin{center}
  \centerline{\includegraphics[width=\columnwidth]{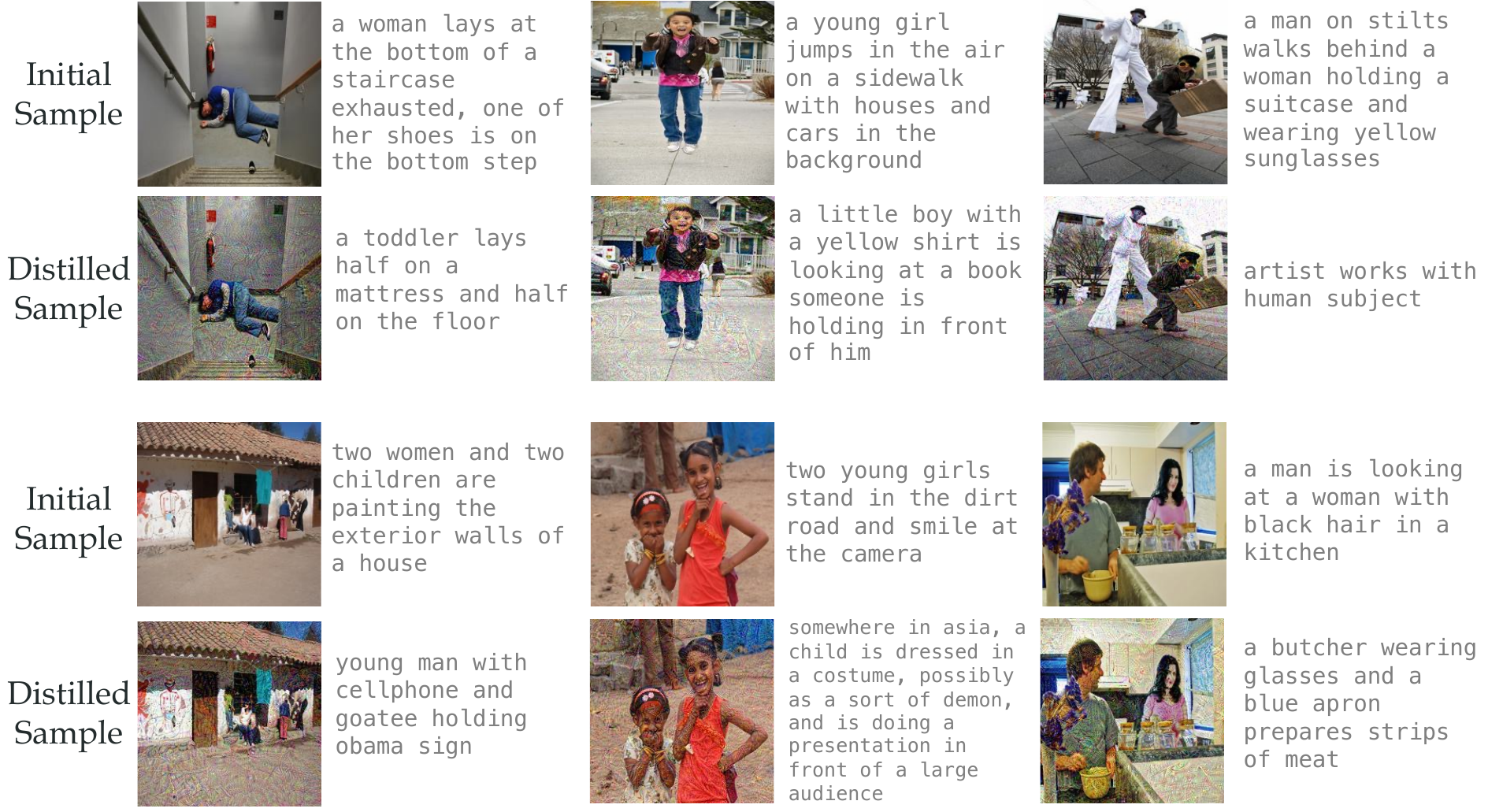}}
  \caption{More examples of synthetic images and retrieval text.}
  \label{fig:more-example}
  \vspace{-15px}
  \end{center}
\end{figure}

In Fig.~\ref{fig:more-example}, we provide more examples of image-text pairs of 200 synthetic pairs for Flickr30k to present the distilled data.

\begin{table*}[t]
\vspace{-3mm}
\caption{Hyperparameters for different experiments.}
\label{tab:param}
\begin{center}
\begin{small}
\begin{sc}
\resizebox{0.5\textwidth}{!}{%
    \begin{tabular}{l|ccc|ccc}
    \toprule
    Dataset & \multicolumn{3}{c|}{Flickr} & \multicolumn{3}{c}{COCO} \\
    Pairs & 100 & 200 & 500  & 100 & 200 & 500   \\
    \midrule
    lr: image       & 100 & 1000 & 1000  & 1000 & 1000 & 5000 \\
    lr: text        & 100 & 1000 & 1000  & 1000 & 1000 & 5000 \\
    lr: lr          & 0.01& 0.01 & 0.01  & 0.01 & 0.01 & 0.01  \\  
    lr: similarity  & 10  & 10   & 100   & 5    & 50   & 500  \\
    initial lr      & 0.1 & 0.1  & 0.1   & 0.1  & 0.1  & 0.1   \\
    batch size      & 20  & 20   & 20    & 20   & 20   & 20    \\
    \midrule
    $\alpha$        &  3 & 1 & 0.01 & 1  & 1  & 1  \\
    rank $r$        & 10 & 5 & 20   & 10 & 20 & 20 \\
    \midrule
    max start epoch &  2  &  2  &  3  &  2  &  2  &  2  \\
    synth steps     &  8  &  8  &  8  &  8  &  8  &  8   \\
    expert epochs   &  1  &  1  &  1  &  1  &  1  &  1   \\
    \bottomrule
    \end{tabular}%
}
\end{sc}
\end{small}
\end{center}
\vspace{-3mm}
\end{table*}

\section{Hyper-parameters}
\label{app-sec:param}

We tune the hyper-parameters and list the values in Tab.~\ref{tab:param}. Many parameters of MTT are directly adopted from previous work~\cite{vl-distill}.

\section{Limitation}

\added{
\begin{enumerate}
    \item \textbf{Distillation of text}. Currently, we are only focusing on learning the synthetic text feature since direct distillation of text tokens is still under investigation and technically not feasible yet. We hope this issue can be addressed in future research.
    \item \textbf{Storage of similarity matrix}. The similarity matrix takes additional storage. Though we have exploited the low-rank method to reduce the memory overhead of LoRS to negligibly linear complexity, there may be a trade-off between storage (the selection of rank $r$) and the performance which complicates the hyperparameter tuning.
    \item \textbf{Loss design}. To use LoRS, the contrastive loss functions should be redesigned and chosen with empirical comparison.
\end{enumerate}
}

\end{document}